\documentclass[letterpaper]{article} 
\usepackage{aaai2026}  
\usepackage{times}  
\usepackage{helvet}  
\usepackage{courier}  
\usepackage[hyphens]{url}  
\usepackage{graphicx} 
\usepackage{natbib}  
\usepackage{caption} 
\usepackage{algorithm}
\usepackage{algorithmic}
\usepackage{algorithm}
\usepackage{algorithmic}

\usepackage{amsmath} 
\usepackage{booktabs}
\usepackage{xcolor}         
\usepackage{graphicx}
\usepackage{subcaption}
\usepackage{colortbl}
\usepackage{pifont}
\usepackage[table]{xcolor}

\usepackage[capitalize]{cleveref}
\crefname{section}{Sec.}{Secs.}
\Crefname{section}{Section}{Sections}
\Crefname{table}{Table}{Tables}

\usepackage{newfloat}
\usepackage{listings}
\DeclareCaptionStyle{ruled}{labelfont=normalfont,labelsep=colon,strut=off} 
\floatstyle{ruled}
\newfloat{listing}{tb}{lst}{}
\floatname{listing}{Listing}
\title{CMMCoT: Enhancing Complex Multi-Image Comprehension via Multi-Modal Chain-of-Thought and Memory Augmentation}
\author{
    Guanghao Zhang\textsuperscript{\rm 1}\equalcontrib,
    Tao Zhong\textsuperscript{\rm 1}\equalcontrib,
    Yan Xia\textsuperscript{\rm 1,2}\equalcontrib,
    Mushui Liu\textsuperscript{\rm 1,2}\equalcontrib,\\
    Zhelun Yu\textsuperscript{\rm 1},
    Haoyuan Li\textsuperscript{\rm 1},
    Wanggui He\textsuperscript{\rm 1},
    Dong She\textsuperscript{\rm 1},
    Yi Wang\textsuperscript{\rm 1,2},
    Hao Jiang\textsuperscript{\rm 1}\thanks{Corresponding authors.}
}
\affiliations{
    \textsuperscript{\rm 1}Alibaba Group, China\\
    \textsuperscript{\rm 2} College of Computer Science and Technology, Zhejiang University, China \\

    \{guanghao.zgh, zt395565\}@taobao.com, \{xiayan.zju, lms\}@zju.edu.cn, yuzhelun.yzl@taobao.com, lihaoyuan@zju.edu.cn, wanggui.hwg@taobao.com, sd0809@mail.ustc.edu.cn, y\_ w@zju.edu.cn,
    aoshu.jh@taobao.com
}

\usepackage{bibentry}

\begin{document}

\maketitle

\begin{abstract}
While previous multimodal slow-thinking methods have demonstrated remarkable success in single-image understanding scenarios, their effectiveness becomes fundamentally constrained when extended to more complex multi-image comprehension tasks. This limitation stems from their predominant reliance on text-based intermediate reasoning processes. While humans, when engaging in sophisticated multi-image analysis, typically perform two complementary cognitive operations: (1) continuous cross-image visual comparison through region-of-interest matching, and (2) dynamic memorization of critical visual concepts throughout the reasoning chain.
Motivated by these observations, we propose the \textbf{Complex Multi-Modal Chain-of-Thought (CMMCoT)} framework, a multi-step reasoning framework that mimics human-like "slow thinking" for multi-image understanding. Our approach incorporates two key innovations: (1) The construction of interleaved multimodal multi-step reasoning chains, which utilize critical visual region tokens, extracted from intermediate reasoning steps, as supervisory signals. This mechanism not only facilitates comprehensive cross-modal understanding but also enhances model interpretability. (2) The introduction of a test-time memory augmentation module that expands the model’s reasoning capacity during inference while preserving parameter efficiency. Furthermore, to facilitate research in this direction, we have curated a novel multi-image slow-thinking dataset. Extensive experiments demonstrate the effectiveness of our model. Code is available at \url{https://github.com/zhangguanghao523/CMMCoT}.

\end{abstract}


\section{Introduction}


Recent years have witnessed the rapid advancement of generative \cite{he2025mars,esser2024scalingrectifiedflowtransformers, wang2025mint, liu2025tfcustom, ma2024followyouremoji, ma2025followcreation, feng2025dit4edit} and multimodal understanding models \cite{wang2025mint, Wang2024Qwen2VLEV}. In particular, multi-modal large language models (MLLMs) have achieved remarkable breakthroughs across various multimodal tasks \cite{liu2024visual, wang2023cogvlm, wang2024qwen2}, such as multimodal recognition, localization, and single- as well as multi-image understanding.
However, current MLLM methods employing the "direct prediction" paradigm to generate answers end-to-end exhibit two critical limitations when confronted with intricate scenarios: \textbf{(a)} They systematically overlook latent evidential features embedded within cross-modal data patterns, resulting in erroneous predictions—even in state-of-the-art models like GPT-4V\cite{yang2023set}, and \textbf{(b)} Their decision-making processes suffer from a pronounced lack of interpretability, undermining their reliability and transparency in critical applications.

The remarkable success of the O1 model has catalyzed growing research interest in chain-of-thought (CoT) reasoning and "slow thinking" mechanisms \cite{snell2024scaling, prystawski2024think}. These mechanisms yield substantial performance enhancements over conventional LLMs and significantly enhance mathematical reasoning and logical deduction capabilities. However, unimodal slow-thinking methodologies exhibit inherent constraints when directly adapted to multimodal domains, primarily due to insufficient cross-modal spatial reasoning capacities for complex scene comprehension.

Existing multimodal CoT approaches predominantly focus on employing external tools to annotate complete textual reasoning chains \cite{ni2024visual, xu2024llava, shao2024visual} or leveraging textual reasoning capabilities to guide visual inference \cite{du2025virgo} while neglecting supervision of visual reasoning processes during training.

Human cognition in multimodal contexts, particularly when processing multi-image compositions, operates through a dual-process cognitive mechanism: (1) parallel processing of linguistic semantics and visual signal interpretation, and (2) synergistic integration of textual deductive reasoning with active visual pattern mining. This cognitive architecture enables the iterative establishment of object-relation correspondences across cross-modal representations, ultimately synthesizing a coherent cross-modal reasoning framework through dynamic interaction between linguistic parsing and visual grounding.
Although previous works like MVoT \cite{li2025imagine} and VoCoT \cite{li2024vocot} have been proposed to mimic the aforementioned process, these methods primarily focus on enhancing the slow reasoning capabilities of models in single-image scenarios, while leaving the more complex multi-image scenarios largely unexplored.

\begin{figure*}[t]
    \centering
    \includegraphics[width=1.0\linewidth]{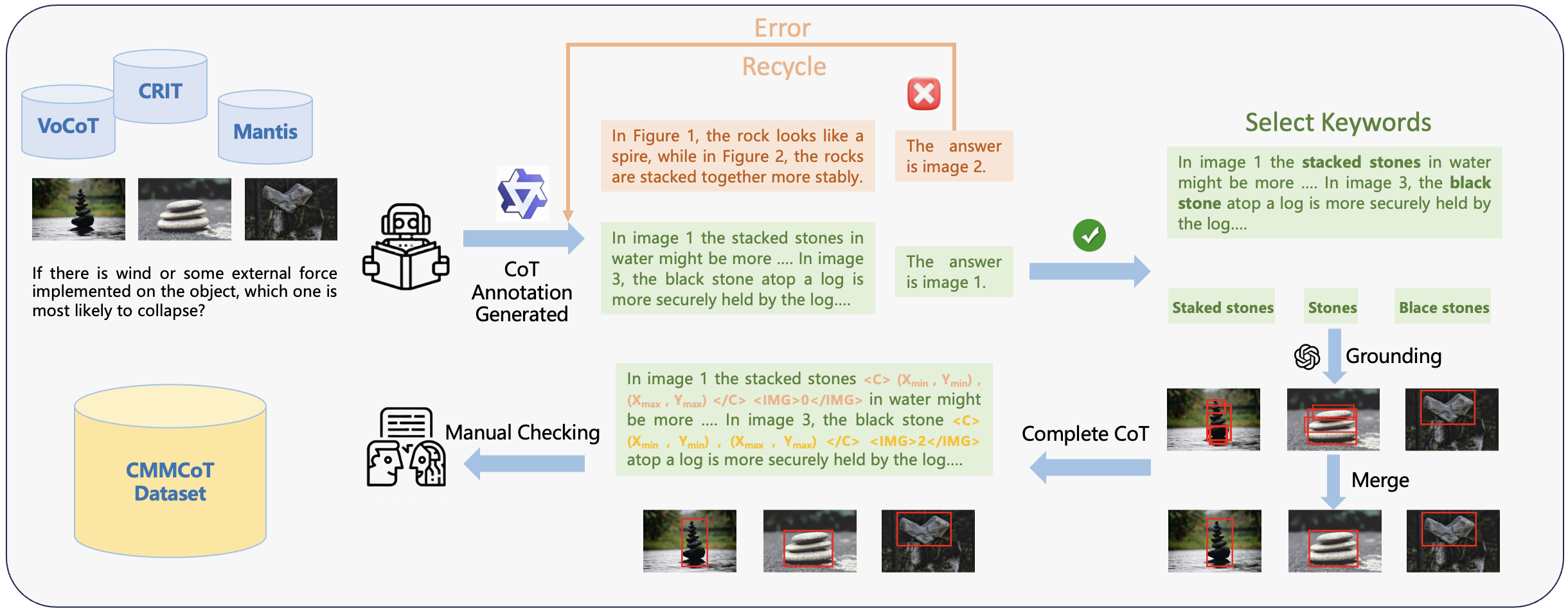}
    \caption{Data generation process of our proposed CMMCoT dataset. The construction of the CMMCoT dataset mainly concerns four parts: the generation of QA rationale chains, the extraction of textual entities, the detection and validation of visual entities, and the spatial fusion and summarization of entity groundings.}
    \label{fig:data}
\end{figure*}

Thus in this paper, we address the challenge of slow thinking in complex multi-image scenarios. However, constructing multi-step reasoning paths for multi-image scenes presents two significant challenges: \textbf{Complexity in Cross-image Visual Concept Tracking:} Unlike single-image scenarios, comprehending multi-image scenes \cite{jiang2024mantis} necessitates the correlation of visual objects across disparate images and the integration of textual information to facilitate comprehensive reasoning. 
\textbf{Enhancement of Model's Inference Capabilities during Testing:} While test-time scaling has shown promise in boosting model reasoning \cite{ni2024visual, xu2024llava} without parameter increases, its effectiveness diminishes in complex multi-image scenarios. Furthermore, simply scaling pre-trained model parameters is reaching a performance ceiling. Thus, exploring alternative methods to enhance model capabilities during testing for multi-image understanding is crucial.

To address the aforementioned challenges, we propose a novel \textbf{Complex Multi-Modal CoT (CMMCoT)} framework that empowers models with slow-thinking capability in multi-image scenarios, significantly enhancing reasoning performance. To overcome the limitations of previous models that are constrained to generating coordinates during reasoning, we devise a novel training-inference paradigm. This paradigm enables the model to extract visual tokens of key objects, grounded by the coordinates mentioned in the reasoning chain, and subsequently predict subsequent reasoning steps and final answers.

Building upon prior works \cite{brown2024large, wang2024augmenting}, we further propose the \textbf{Retrieval-based Image Feature Reasoning Enhancement Module (RIFREM)} module to enable comprehensive cross-modal deliberation during testing. Specifically, this module stores the Key and Value pairs of multi-image input tokens obtained after each decoder layer in a dedicated memory bank. When decoding specific coordinates, the model retrieves corresponding visual tokens using image IDs and coordinates, then computes attention between the query vectors (derived from subgraph token sequences processed through each decoder layer) and the stored multi-image Key/Value pairs in the memory bank.

Furthermore, considering that existing works lack datasets meeting our requirements, we constructed a new dataset named \textbf{CMMCoT-260k}, which constitutes an innovative benchmark specifically designed for complex multi-image multimodal tasks, comprising 260,000 meticulously curated data instances. Distinct from conventional datasets, CMMCoT-260k's uniqueness resides in its incorporation of explicit reasoning chains within each instance. These structured reasoning pathways not only facilitate deep semantic parsing of hybrid text-image data but also integrate spatial coordinates and entity-specific imagery, thereby enabling hierarchical reasoning analysis. 

Extensive experiments conducted on both multi-image and single-image benchmarks demonstrate the effectiveness of our model. Moreover, visualization experiments have also demonstrated that compared to the traditional GPT-4o and Qwen2.5-VL, our model not only improves the accuracy of responses but also significantly enhances the interpretability of the intermediate reasoning process, making it more accessible for human understanding.

\section{Dataset Construction}
Building upon multiple datasets including GRIT \cite{peng2023kosmos}, Flickr30k-Entities \cite{plummer2015flickr30k}, VoCoT \cite{li2024vocot}, and MANTIS \cite{jiang2024mantis}, we have constructed a complex multi-image, multi-modal Chain-of-Thought (CoT) dataset, referred to as the \textbf{CMMCoT-260K} dataset. It consists of 260,000 instances, encompassing four distinct task types: {Caption}, {Co-reference}, {Comparison}, and {Reason}.

For the {Caption} and {Co-reference} tasks, we employed a straightforward data integration pipeline. In contrast, for the more challenging {Comparison} and {Reason} tasks, we designed a more complex data processing pipeline that includes the following steps:

\begin{itemize}
\item \textbf{Constructing QA Rationale Chains:} We generate CoT annotations through a fully automated pipeline based on the methodology from \cite{zelikman2024star}. Initially, GPT-4o is used to generate a preliminary reasoning chain and answer from the question. If the answer is correct, the chain is retained. If incorrect, the question is paired with the pre-annotated correct answer and re-fed into GPT-4o to generate a refined rationale. This process uses answer correctness as a quality filter to ensure the reliability of the generated rationales.

\item \textbf{Entity Extraction:} The Qwen3-235B-A22B model is utilized to extract textual entities from QA dialogues, providing a foundation for subsequent entity localization and relationship summarization.

\item \textbf{Entity Detection:} To ensure localization accuracy, we adopt a two-stage validation mechanism. First, the Qwen-VL-max model generates initial bounding boxes for entities. Subsequently, GPT-4o calculates the Intersection over Union (IoU) for these boxes, and only samples with an IoU $\geq$ 0.9 are retained to ensure precision.

\item \textbf{Entity Relationship Summarization:} When a single entity corresponds to multiple bounding boxes across images or detections, we apply spatial fusion. The smallest top-left and largest bottom-right coordinates from all relevant boxes are used to create a single, unified bounding box. This approach simplifies the CoT structure while maintaining complete entity coverage, enhancing the clarity and logical coherence of the reasoning chain.
\end{itemize}

In summary, through this pipeline, we have successfully transformed multi-image QA datasets into a comprehensive and powerful multi-image, multi-modal CMMCoT dataset, establishing an essential foundation for sophisticated multi-modal reasoning systems.

\begin{figure*}[t]
    \centering
    \includegraphics[width=1.0\linewidth]{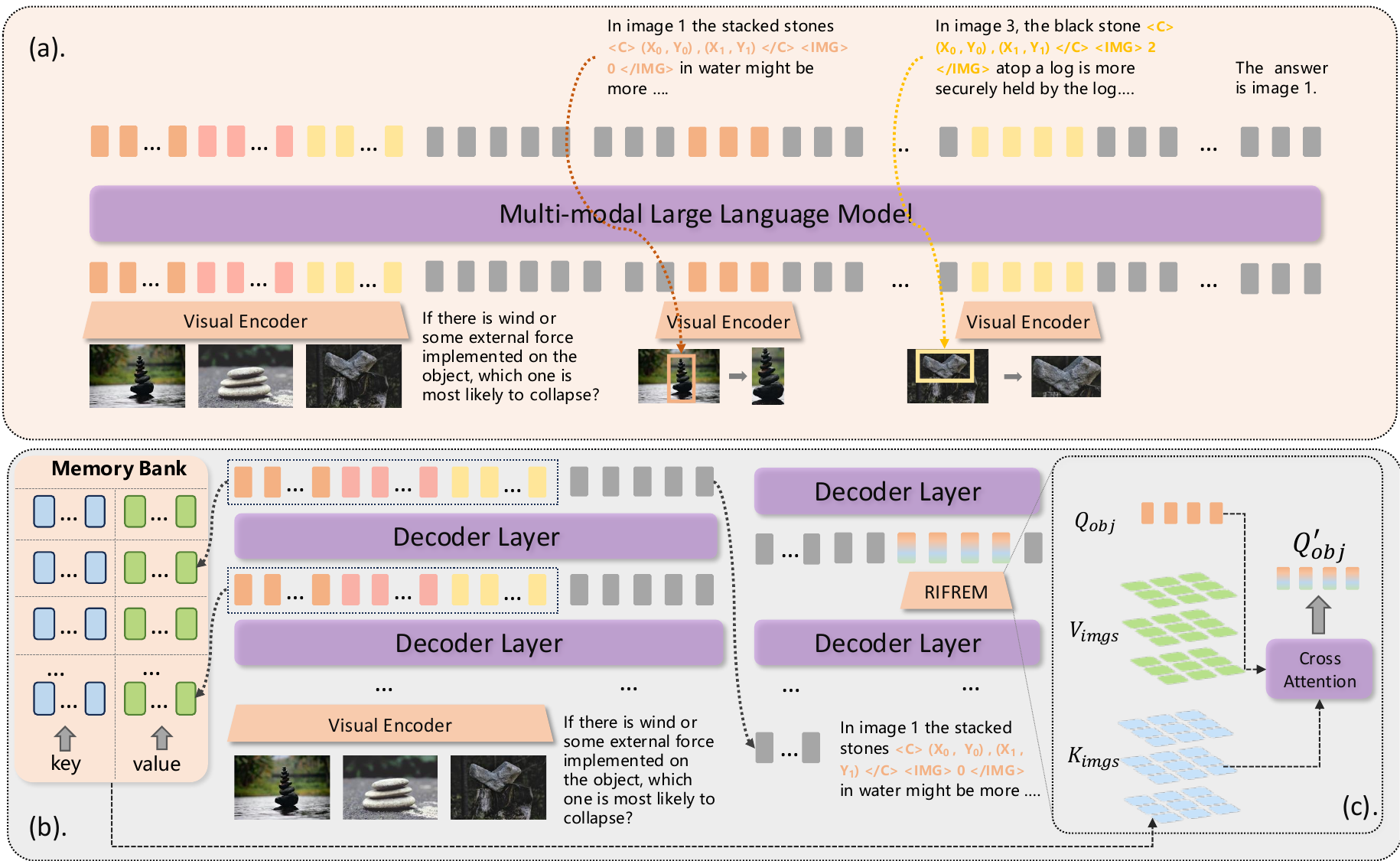}
    \caption{\textbf{Illustration of the overall framework of CMMCoT.} Part (a) depicts the training structure, with the input and output shown below and above the model, respectively. Part (b) presents the inference structure of the model, where the memory bank stores the K and V for each layer of the input images. The RIFREM module is integrated between different decoder layers during inference. Part (c) represents the detailed structure of the RIFREM module.}
    \label{fig:model}
    \vspace{-3mm}
\end{figure*}

\section{Methods}

\subsection{Baseline}
Along with the proposed dataset, we also develop a novel multi-modal framework named CMMCoT, which employs Qwen2-VL \cite{wang2024qwen2} as the baseline model. The primary reason for selecting Qwen2-VL is its trained Vision Transformer (ViT) that supports Naive Dynamic Resolution, enabling the processing of images at arbitrary resolutions and dynamically converting them into a variable number of visual tokens. Furthermore, Qwen2-VL introduces Multimodal Rotary Position Embedding (M-RoPE), which effectively models the positional information of multimodal inputs. This approach reduces the positional ID values of images and videos, allowing the model to extrapolate to longer sequences during inference.

\subsection{Multimodal Sequence Representation} CMMCoT represents complex multi-image and textual data in an interleaved visual-textual format. For complex multi-image tasks, it is essential to compare and contrast the associations and differences among different images, between different entities within the same image, and between different entities across different images during the reasoning process. Therefore, we introduce special image index tokens to refer to specific input images, formatted as \textless{}IMG\textgreater{}0\textless{}\text{/IMG}\textgreater{}. Here, \textless{}IMG\textgreater{} and \textless{}\text{/IMG}\textgreater{} are special tokens indicating the start and end of an image index. This indexing logic can be extended to accommodate more images.

Additionally, we represent different entities using coordinates and visual tokens. The coordinate format is similar to \textless{}\textbar{}box\_start\textbar{}\textgreater{}(\textless{}$x_0$\textgreater{}, \textless{}$y_0$\textgreater{}) , (\textless{}$x_1$\textgreater{}, \textless{}$y_1$\textgreater{})\textless{}\textbar{}box\_end\textbar{}\textgreater{}, where \textless{}\textbar{}box\_start\textbar{}\textgreater{} and \textless{}\textbar{}box\_end\textbar{}\textgreater{} are special markers indicating the beginning and end of coordinate information. We use bounding boxes (\textless{}$x_0$\textgreater{}, \textless{}$y_0$\textgreater{}) and (\textless{}$x_1$\textgreater{}, \textless{}$y_1$\textgreater{}) as the entity coordinates, with x and y normalized relative to the image dimensions, ranging from 0 to 1000. The coordinate values are tokenized and embedded as textual data.

For visual information pertaining to multiple images and entities, we use special markers \textless{}\textbar{}vision\_start\textbar{}\textgreater{} and \textless{}\textbar{}vision\_end\textbar{}\textgreater{} to denote the start and end positions of visual content. During the training process, we obtain entity images based on their coordinates and image indices and encode each entity image using a visual encoder. When encoding the entity images through the visual encoder, we limit the minimum resolution of the entities to 512 pixels, which allows us to extract more detailed features from the entity images \cite{li2024mini}.

\textbf{Training:}
To achieve better performance on complex multi-image tasks without compromising single-image performance, our CMMCoT method employs a two-stage training strategy:
\begin{itemize}
    \item Stage 1: Multi-Image Training. We begin by training on our self-constructed CMMCoT-260k dataset, which is specifically designed for complex multi-image tasks. This phase aims to enable the model to handle intricate tasks involving multiple images.
    \item Stage 2: Mixed Training. In this phase, the CMMCoT-260k dataset is mixed with a general dataset at a 1:1 sampling ratio for training. The goal of this strategy is to significantly alleviate catastrophic forgetting caused by multi-image tasks while retaining the model's general visual understanding capabilities.
\end{itemize}

The training objective function is to minimize the loss in predicting the next token.
During training, for the CMMCoT-260k dataset, the model input consists of multiple images and related questions, and the output is an answer that encompasses the reasoning process, entity coordinates, and entity images. We append the prompt \textit{"Please answer the question with reasoning and identify key objects."} after the question to enhance the model's reasoning ability. Entity images are extracted based on the provided entity coordinates and image indices and are encoded using a visual encoder. During training, the loss is calculated only on the text, coordinates, and special tokens; the entity image part does not contribute to the loss. For the general dataset, the training process follows the standard multimodal training procedure.

\textbf{Inference:} 
During the inference process, we also apply the prompt \textit{"Please answer the question with reasoning and identify key objects."} to stimulate the model's reasoning capabilities in both single-image and multi-image tasks. The output of text and coordinates is consistent with that of standard multimodal models. However, when the \textless{}\text{/IMG}\textgreater{} token is predicted, the entity image is extracted based on the image index and coordinates. Subsequently, the proposed RIFREM module is utilized to extract relevant features between the entity and the input multiple images, thereby enriching the entity features and continuing to reason over the subsequent tokens.

\begin{table*}[t]
\centering
\resizebox{\textwidth}{!}{  
\begin{tabular}{l c c c c c c c}
\toprule
\textbf{Model} & \textbf{Params} & \textbf{BLINK} & \textbf{Mantis} & \textbf{NLVR2} & \textbf{MVBench} & \textbf{Q-Bench} & \textbf{Avg} \\
\midrule

LLaVA-v1.5 & 7B & 37.1 & 41.9 & 52.1 & 36.0 & 53.9 & 44.2 \\
LLaVA-v1.6 & 7B & 39.6 & 45.6 & 58.9 & 40.9 & 58.9 & 48.8 \\
LLaVA-v1.5-MIA-DPO & 7B & 42.9 & 44.2 & 54.2 & 39.5 & -- & -- \\
LLaVA-OV & 7B & 48.2 & 64.2 & 89.4 & 56.7 & 74.5 & 66.6 \\

Mantis-Idefics2 & 8B & 49.1 & 57.1 & 89.7 & 51.4 & 75.3 & 64.5 \\
InternVL3 & 8B & 55.5 & 70.1 & 88.5 & 75.4 & 75.9 & 73.1 \\

\rowcolor{gray!20}
Qwen2-VL & 7B & 51.8 & 64.8 & 84.0 & 50.5 & 71.2 & 64.5 \\
\rowcolor{gray!20}
Qwen2-VL (Ours) & 7B & \textbf{52.3} & \textbf{70.9} & \textbf{88.7} & \textbf{51.3} & \textbf{72.2} & \textbf{67.1} \\

\rowcolor{gray!35}
Qwen2.5-VL & 7B & 55.3 & 69.8 & 88.3 & 74.7 & 77.7 & 73.2 \\
\rowcolor{gray!35}
Qwen2.5-VL (Ours) & 7B & \textbf{56.8} & \textbf{72.2} & \textbf{89.9} & \textbf{75.8} & \textbf{78.5} & \textbf{74.6} \\

\midrule

InternVL3 & 2B & 50.3 & 65.9 & 85.4 & 70.4 & 71.5 & 68.7 \\

\rowcolor{gray!20}
Qwen2-VL & 2B & 43.6 & 37.7 & 74.1 & 37.5 & 58.5 & 50.3 \\
\rowcolor{gray!20}
Qwen2-VL (Ours) & 2B & \textbf{44.8} & \textbf{39.8} & \textbf{76.9} & \textbf{40.1} & \textbf{61.7} & \textbf{52.7} \\

\rowcolor{gray!35}
Qwen2.5-VL & 3B & 49.1 & 62.7 & 86.2 & 71.3 & 74.9 & 68.8 \\
\rowcolor{gray!35}
Qwen2.5-VL (Ours) & 3B & \textbf{51.4} & \textbf{68.5} & \textbf{88.9} & \textbf{73.1} & \textbf{75.2} & \textbf{71.4} \\

\bottomrule
\end{tabular}
}
\caption{Performance comparison of SOTA models on multi-image benchmarks. Rows in \colorbox{gray!20}{\textcolor{black}{light gray background}} and \colorbox{gray!35}{\textcolor{black}{darker gray background}} represent Qwen2-VL and Qwen2.5-VL variants, respectively. Best results per benchmark are in \textbf{bold}.}
\vspace{-3mm}
\label{tab:model_performance}
\end{table*}

\subsection{RIFREM}
Previous studies \cite{snell2024scaling, prystawski2024think} have demonstrated that expanding model computational capacity during inference can significantly enhance performance metrics. For multi-image tasks, a natural approach involves comparing critical visual regions with relevant areas in other images through joint reasoning when identifying pivotal visual attention zones.

To enable cross-image feature mining among entity-related images during inference, we propose the \textbf{Retrieval-based Image Feature Reasoning Enhancement Module (RIFREM).} This framework treats entity image features as queries (Q) while utilizing multi-image features as keys (K) and values (V) to retrieve relevant visual cues from input multi-image candidates. Specifically, we maintain a memory bank $\mathcal{M}$ to store the key-value pairs of multi-image sequences from each decoder layer during multi-image input processing. The memory bank can be formulated as:

\begin{equation}
    \mathcal{M} = \{ (K_l^{(i)}, V_l^{(i)}) \}_{l=1, 2, ..., L; i=1, 2, ..., N}
\end{equation}
where $K_l^{(i)}, V_l^{(i)}$ represents the keys and values from the $l$-th decoder layer of the $i$-th multi-image sequence, $l$ is the number of decoder layers, and $N$ is the number of multi-image sequences processed.

During CoT prediction, when encountering the \textless{}\text{/IMG}\textgreater{} token in the reasoning process, we inject entity image tokens into decoder layers and extract their query vectors based on position IDs. These queries then engage in cross-attention with corresponding key-value pairs retrieved from the memory bank. The attention mechanism can be formally described as scaled dot-product attention:

\begin{equation}
    Q' = \text{Att}(Q, K_{\mathcal{M}}, V_{\mathcal{M}}) = \text{softmax}\left(\frac{QK_{\mathcal{M}}^T}{\sqrt{d_k}}\right)V_{\mathcal{M}}
\end{equation}
where $K_{\mathcal{M}}$ and $V_{\mathcal{M}}$ are the keys and values retrieved from the memory bank $\mathcal{M}$ corresponding to the current query $Q$, and $d_k$ is the dimension of the keys. 
These refined $Q'$ subsequently propagate through subsequent reasoning stages. Our ablation studies reveal that incorporating RIFREM modules across different decoder layers introduces non-trivial latency overheads, prompting systematic comparisons between performance gains and computational costs.

\begin{table*}[t]
\centering
\resizebox{\textwidth}{!}{
\small
\begin{tabular}{lcrrrrrrrr}
\toprule
\textbf{Models} & \textbf{Params} & \textbf{MMMU} & \textbf{MMStar}& \textbf{SQA} & \textbf{RealWorldQA} & \textbf{MME} & \textbf{POPE} & \textbf{HallBench} & \textbf{Avg} \\
\midrule
LLaVA-v1.5 & 7B & 35.1 & 32.9 & 66.6 & 48.9 & 58.4 & 78.0 & 35.8 & 50.8 \\
LLaVA-v1.6 & 7B & 35.8 & 37.6 & 69.7 & 60.8 & 63.6 & 86.1 & 30.3 & 54.8 \\
LLaVA-MIA-DPO & 7B & 36.3 & 32.9 & 67.6 & -- & -- & 87.2 & -- & -- \\
LLaVA-OV & 7B & 48.8 & 61.7 & 94.8 & 66.3 & 71.4 & 87.1 & 48.3 & 68.4 \\
Mantis-Idefics2 & 8B & 40.1 & 46.4 & 77.3 & 59.3 & 64.5 & 82.8 & 56.7 & 61.0 \\
InternVL3       & 8B & 62.7 & 68.7 & 97.9 & 71.4 & 86.5 & 90.4 & 49.0 & 75.2 \\
\rowcolor{gray!20} Qwen2-VL & 7B & \textbf{51.6} & 57.7 & 85.5 & 63.3 & \textbf{82.7}& \textbf{88.4}& 50.6 & 68.5 \\
\rowcolor{gray!20} Qwen2-VL (Ours) & 7B & 48.2 & \textbf{58.9} & \textbf{88.9} & \textbf{65.0} & 81.5 & 86.3 & \textbf{60.5} & \textbf{69.9} \\
\rowcolor{gray!35} Qwen2.5-VL      & 7B & \textbf{58.6} & 63.9 & 89.0 & 68.4 & 82.6 & \textbf{85.9} & 51.9 & 71.4 \\
\rowcolor{gray!35} Qwen2.5-VL (Ours) & 7B & 57.5 & \textbf{66.4} & \textbf{96.8} & \textbf{71.6} & \textbf{83.5} & 89.2 & \textbf{63.6} & \textbf{75.5} \\
\midrule
InternVL3       & 2B & 48.6 & 61.1 & 95.8 & 65.1 & 78.1 & \textbf{90.1} & 41.9 & 68.7 \\
\rowcolor{gray!20} Qwen2-VL & 2B & \textbf{41.1}& 48.0 & 73.2 & 62.3 & \textbf{66.8}& \textbf{85.8}& 41.7 & 59.8 \\
\rowcolor{gray!20} Qwen2-VL (Ours) & 2B & 39.4 & \textbf{49.2}& \textbf{78.6}& \textbf{62.6}& 65.7 & 84.0 & \textbf{52.5}& \textbf{61.7}\\
\rowcolor{gray!35} Qwen2.5-VL      & 3B & \textbf{53.1} & 55.8 & 81.4 & 65.5 & 78.6 & 85.9 & 46.6 & 66.7 \\
\rowcolor{gray!35} Qwen2.5-VL (Ours) & 3B & 53.0 & \textbf{57.3} & \textbf{88.6} & \textbf{ 67.2} & \textbf{79.8} & \textbf{87.1} & \textbf{48.6} & \textbf{68.8} \\
\bottomrule
\end{tabular}
\vspace{0.2cm}
\begin{minipage}{\textwidth}
\raggedright\footnotesize
\end{minipage}
}
\caption{Performance Comparison of SOTA Models across Single-image Benchmarks. The Qwen2-VL series are highlighted with \colorbox{gray!20}{\textcolor{black}{light gray background}} and Qwen2.5-VL series with \colorbox{gray!35}{\textcolor{black}{darker gray background}}, while other models use white background. Bold numbers indicate the best performance for each benchmark.}
\label{tab:model_single_image_benchmark}
\end{table*}

\begin{figure*}[!t]
    \centering
    \includegraphics[width=1\linewidth]{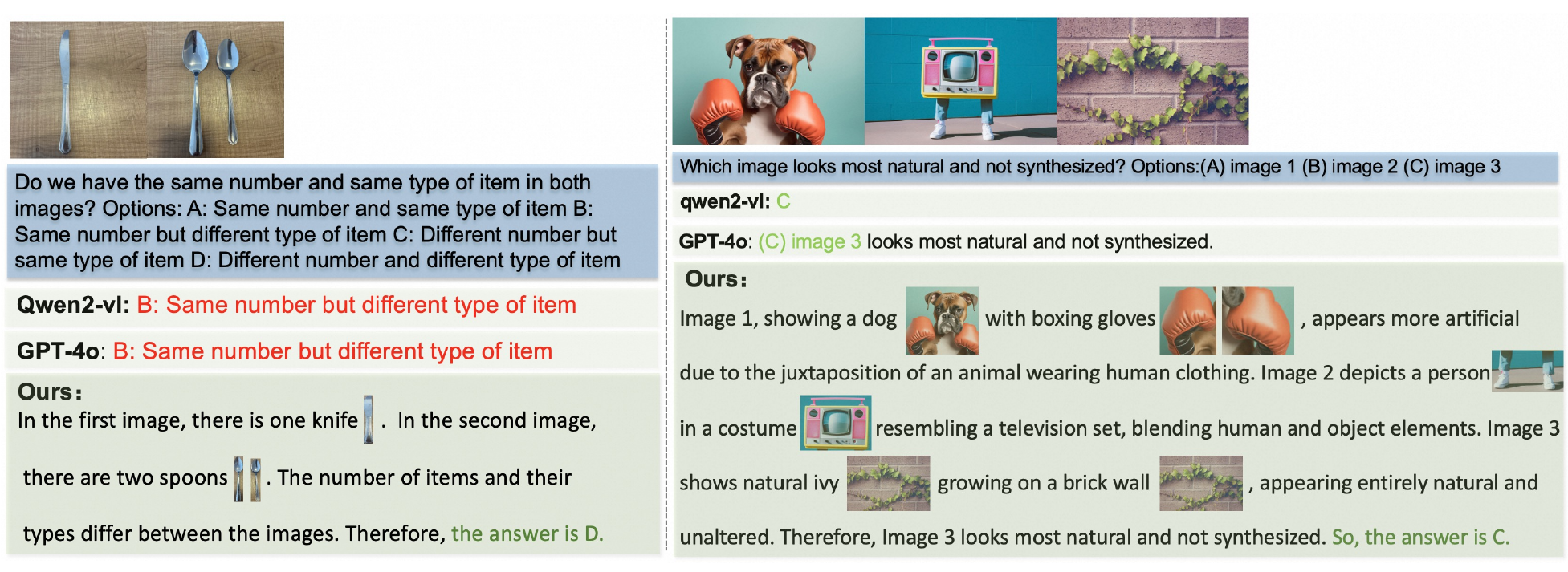}
    \caption{Visualization results of our CMMCoT task to illustrate the difference between our model and previous methods.}
    \label{fig:demo2}
\end{figure*}

\begin{table*}[t]
  \centering
  \setlength{\tabcolsep}{8pt}
  \begin{tabular}{ccc|cccccc}
    \toprule
    \textbf{Grounding} & \textbf{MCoT} & \textbf{RIFREM} &
    \textbf{BLINK} & \textbf{Mantis} & \textbf{NLVR2} & \textbf{MVBench} & \textbf{Q-Bench} & \textbf{Avg} \\
    \midrule
    \ding{56} & \ding{56} & \ding{56} & 55.3 & 69.8 & 88.3 & 74.7 & 77.7 & 73.2 \\
    \ding{51} & \ding{56} & \ding{56} & 55.2 & 70.4 & 88.7 & 74.5 & 77.9 & 73.3 \\
    \ding{51} & \ding{51} & \ding{56} & \textbf{57.1} & 71.6 & 89.4 & 75.4 & \textbf{78.7} & 74.4 \\
    \ding{51} & \ding{51} & \ding{51} & \textbf{56.8} & \textbf{72.2} & \textbf{89.9} & \textbf{75.8} & 78.5 & \textbf{74.6} \\
    \bottomrule
  \end{tabular}
  \caption{Performance (\%) of Qwen2.5-VL-7B variants across benchmarks with different module combinations}
  \label{tab:model_benchmarks}
\end{table*}

\section{Experiments}
\paragraph{Evaluation Benchmarks}
\textbf{NLVR2} \cite{suhr2018corpus} evaluates visual-textual reasoning through cross-image logical analysis, presenting paired images with declarative statements for propositional verification. It employs constrained selection on the standard test-public partition.
\textbf{Qbench} \cite{wu2023q} assesses multimodal models' critical analysis of benchmarking parameters, focusing on visual perception in quality assessment tasks. Our experiment employs its Qbench2-A1-dev subset with multiple-choice comparative visual analysis.
\textbf{Mantis-Eval} contains 217 multi-image analysis tasks across conceptual domains (dimensional assessment, mass estimation). Manually curated from web-sourced images, it combines closed and open-response formats evaluated on standardized test partitions.
\textbf{BLINK} \cite{fu2025blink} tests rapid visual cognition (depth relationships, feature matching, digital forensics, spatial-temporal reasoning). It includes multi-image perceptual similarity tasks, measured through standardized validation protocols.
\textbf{MVBench} \cite{li2024mvbench} evaluates temporal reasoning in video comprehension through 20 tasks analyzing frame dynamics. Using 8-frame sampling per video, results are quantified via standardized test-set measurements.
\textbf{Single Image Benchmarks:} We also conduct experiments on single-image tasks to further demonstrate the effectiveness of our proposed model: MMMU \cite{yue2024mmmu}, MMStar \cite{chen2024we}, SQA \cite{lu2022learn}, RealWorldQA, MME \cite{yin2023survey}, POPE \cite{li2023evaluating}, HallBench \cite{guan2023hallusionbench}.

\paragraph{Experiments Setting}
We conduct supervised fine-tuning (SFT) based on the Qwen2.5-VL-7B architecture. The optimization process employs AdamW with $\beta=0.95$ and a weight decay of 0.1, coupled with a cosine learning rate scheduler. The training framework utilizes the DeepSpeed ZeRO-3 strategy for efficient parameter optimization. The training regimen consists of two distinct phases:
Stage 1: Initial learning rate of  1\text{e}{-5}  with 2 training epochs.  Stage 2: Reduced learning rate of  1\text{e}{-6} with 1 training epoch at batch size 256. This two-phase approach enables progressive refinement of model parameters, where the first stage establishes coarse-grained feature representations and the second stage performs fine-grained adjustment with reduced stochasticity through smaller learning rates and larger batch sizes.

\paragraph{Compared Methods}
To evaluate and compare the effectiveness of multi-image models, we conducted comprehensive benchmarking on several state-of-the-art architectures: LLaVA-v1.5 \cite{liu2024improved}, LLaVA-v1.6 \cite{li2024llava}, LLaVA-v1.5-MIA-DPO \cite{liu2024mia}, Mantis-Idefics2 \cite{jiang2024mantis}, LLaVA-OV \cite{li2024llava}, Qwen2-VL \cite{Wang2024Qwen2VLEV},  InternVL3 \cite{zhu2025internvl3} and Qwen2.5-VL \cite{Bai2025Qwen25VLTR}. Our evaluation framework simultaneously addresses both multi-image and single-image tasks, enabling comparative analysis of model performance across different input configurations.

\paragraph{Performance Evaluation}
\textbf{Multi-image performance:} The experimental results across multiple diagnostic benchmarks demonstrate the consistent superiority of our model across both 7B and 3B parameter configurations. As shown in Tab.~\ref{tab:model_performance}, our 7B variant achieves state-of-the-art performance on Mantis-Eval (72.2), MVBench(75.8), Q-Bench(78.5), and competitive results on NLVR2 (89.9) and BLINK (52.3), outperforming comparable 7B models like Qwen2.5-VL by 1.4 average points. Notably, the scaled-down 3B version maintains robust capabilities, surpassing its parameter-matched counterpart Qwen2.5-VL-3B by 2.6 average points while exhibiting particular strengths in temporal reasoning (MVBench: 73.1) and visual-textual verification (NLVR2: 88.9). This dual-scale effectiveness suggests our architecture successfully balances model capacity with cross-modal alignment efficiency, preserving critical visual reasoning abilities even in parameter-constrained scenarios. The performance parity across distinct evaluation paradigms from multi-image logical analysis (Mantis) to perceptual similarity judgments (BLINK) validates our approach's adaptability in handling diverse visual reasoning tasks through enhanced feature interaction mechanisms. 

\textbf{Single-image performance:} Similarly, as shown in Tab.~\ref{tab:model_single_image_benchmark} in single-image datasets, our model consistently achieves superior performance across different scales compared to prior architectures, demonstrating that the proposed multi-modal hybrid sequential chain reasoning strategy effectively enhances comprehension capabilities for single-image tasks. This cross-scale effectiveness confirms the generalizability of our methodology beyond multi-image analytical scenarios, particularly in improving fundamental visual-semantic alignment through structured reasoning pathways.

\subsection{Quantitive Analysis}
To further evaluate how multimodal hybrid sequence CoT enhances model understanding in complex multi-image scenarios, we visualize representative examples in Fig.~\ref{fig:demo2}. The left example in Fig.~\ref{fig:demo2} highlights our model's superiority in object localization and quantitative reasoning. While other models fail to correctly detect or count cutlery items (e.g., knives and spoons), our model accurately identifies their positions and quantities, demonstrating robust counting ability even in ambiguous visual contexts. The right example illustrates our model's advantage in anomaly detection and logical reasoning. Although alternative models may provide correct answers, our method not only identifies commonsense violations in the visual input but also offers explicit causal interpretations. This capacity for contextual inconsistency detection underscores the enhanced reasoning capabilities enabled by our approach.

\subsection{Ablation Study}
\textbf{Impact of Different Modules.} We conducted a series of ablation studies to examine the contributions of individual components, as shown in Table~\ref{tab:model_benchmarks}. We first analyzed the effect of using only grounding-based information during training. This setting yielded a modest 0.1-point improvement in average score, suggesting limited benefit when used in isolation. Next, we investigated the impact of combining grounding with entity images. This joint supervision led to a notable performance gain, improving the average score from 73.2 to 74.4. The result highlights the synergy between grounding cues and visual features in enhancing multimodal representation learning. We further evaluated the RIFREM module applied during inference. Its integration provided an additional boost, raising the average score to 74.6 by exploiting cross-image feature relationships. Together, these experiments clarify the individual and combined contributions of each module. They underscore the importance of coordinated feature utilization—across both training and inference—for optimizing performance in multimodal tasks.

\begin{figure}[!t]
    \centering
    \includegraphics[width=0.9\linewidth]{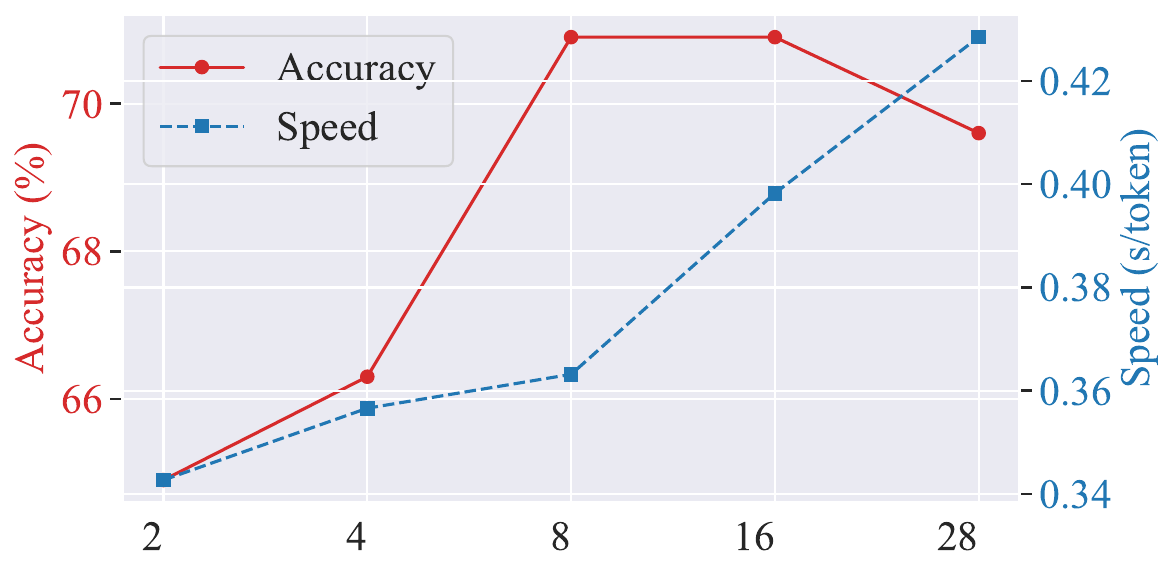}
    \caption{\textbf{Ablation study on the number of RIFREM module layers:} The red line represents the performance effects, while the blue line indicates the impact on latency.}
    \label{fig:line_plot}
\end{figure}

\noindent
\textbf{Impacts of RIFREM.}
We evaluate the effect of integrating the RIFREM module at different network depths during inference, balancing performance gains and latency overhead (Figure~\ref{fig:line_plot}). Five configurations were tested, from shallow (Group 1: layers 0 and 27) to full-depth integration (Group 5: all layers). Results show that applying RIFREM only at the first and last layers degrades performance, likely due to disrupted information flow. In contrast, progressive integration across the network (Groups 2–5) enables iterative refinement and better feature consistency. Group 3 (8 evenly distributed layers) achieves the best trade-off between accuracy and efficiency, and is used in all subsequent experiments.

\section{Conclusion}
In this paper, we propose CMMCoT, a framework enabling systematic slow-thinking for multi-modal large language models (MLLMs) in complex multi-image scenarios. By integrating coordinate-guided visual token extraction and a Retrieval-based Image Feature Reasoning Enhancement Module (RIFREM), CMMCoT mitigates error accumulation and enhances cross-image visual concept tracking. Evaluations across six benchmarks demonstrate state-of-the-art performance, with RIFREM facilitating dynamic deliberation over multi-image contexts during inference. Case studies highlight improved accuracy and interpretability over GPT-4V. This work underscores the necessity of structured reasoning mechanisms for advancing MLLMs, paving the way for future research in dynamic multi-modal reasoning and memory-efficient architectures.

\section{Acknowledgments}
This research was supported in part by Zhejiang Provincial Natural Science Foundation of China under Grant LD24F020016, the Key R\&D  Program of Zhejiang Province 2025C01075, 2023C01043, the National Natural Science Foundation of China under Grant 62576313, and Alibaba Group through Alibaba Research Intern Program.

\bibliography{aaai2026}

@article{yang2023set,
  title={Set-of-mark prompting unleashes extraordinary visual grounding in gpt-4v},
  author={Yang, Jianwei and Zhang, Hao and Li, Feng and Zou, Xueyan and Li, Chunyuan and Gao, Jianfeng},
  journal={arXiv preprint arXiv:2310.11441},
  year={2023}
}

@article{li2025imagine,
  title={Imagine while Reasoning in Space: Multimodal Visualization-of-Thought},
  author={Li, Chengzu and Wu, Wenshan and Zhang, Huanyu and Xia, Yan and Mao, Shaoguang and Dong, Li and Vuli{\'c}, Ivan and Wei, Furu},
  journal={arXiv preprint arXiv:2501.07542},
  year={2025}
}

@article{li2024vocot,
  title={VoCoT: Unleashing Visually Grounded Multi-Step Reasoning in Large Multi-Modal Models},
  author={Li, Zejun and Luo, Ruipu and Zhang, Jiwen and Qiu, Minghui and Wei, Zhongyu},
  journal={arXiv preprint arXiv:2405.16919},
  year={2024}
}

@inproceedings{wang2024augmenting,
  title={Augmenting language models with long-term memory},
  author={Wang, Weizhi and Dong, Li and Cheng, Hao and Liu, Xiaodong and Yan, Xifeng and Gao, Jianfeng and Wei, Furu},
  booktitle={NeurIPS},
  year={2024}
}

@article{brown2024large,
  title={Large language monkeys: Scaling inference compute with repeated sampling},
  author={Brown, Bradley and Juravsky, Jordan and Ehrlich, Ryan and Clark, Ronald and Le, Quoc V and R{\'e}, Christopher and Mirhoseini, Azalia},
  journal={arXiv preprint arXiv:2407.21787},
  year={2024}
}

@article{ni2024visual,
  title={Visual-o1: Understanding ambiguous instructions via multi-modal multi-turn chain-of-thoughts reasoning},
  author={Ni, Minheng and Fan, Yutao and Zhang, Lei and Zuo, Wangmeng},
  journal={arXiv preprint arXiv:2410.03321},
  year={2024}
}

@article{xu2024llava,
  title={LLaVA-o1: Let Vision Language Models Reason Step-by-Step},
  author={Xu, Guowei and Jin, Peng and Hao, Li and Song, Yibing and Sun, Lichao and Yuan, Li},
  journal={arXiv preprint arXiv:2411.10440},
  year={2024}
}

@inproceedings{shao2024visual,
  title={Visual cot: Advancing multi-modal language models with a comprehensive dataset and benchmark for chain-of-thought reasoning},
  author={Shao, Hao and Qian, Shengju and Xiao, Han and Song, Guanglu and Zong, Zhuofan and Wang, Letian and Liu, Yu and Li, Hongsheng},
  booktitle={The Thirty-eight Conference on Neural Information Processing Systems Datasets and Benchmarks Track},
  year={2024}
}

@article{du2025virgo,
  title={Virgo: A Preliminary Exploration on Reproducing o1-like MLLM},
  author={Du, Yifan and Liu, Zikang and Li, Yifan and Zhao, Wayne Xin and Huo, Yuqi and Wang, Bingning and Chen, Weipeng and Liu, Zheng and Wang, Zhongyuan and Wen, Ji-Rong},
  journal={arXiv preprint arXiv:2501.01904},
  year={2025}
}

@article{peng2023kosmos,
  title={Kosmos-2: Grounding multimodal large language models to the world},
  author={Peng, Zhiliang and Wang, Wenhui and Dong, Li and Hao, Yaru and Huang, Shaohan and Ma, Shuming and Wei, Furu},
  journal={arXiv preprint arXiv:2306.14824},
  year={2023}
}

@inproceedings{plummer2015flickr30k,
  title={Flickr30k entities: Collecting region-to-phrase correspondences for richer image-to-sentence models},
  author={Plummer, Bryan A and Wang, Liwei and Cervantes, Chris M and Caicedo, Juan C and Hockenmaier, Julia and Lazebnik, Svetlana},
  booktitle={Proceedings of the IEEE international conference on computer vision},
  pages={2641--2649},
  year={2015}
}

@article{jiang2024mantis,
  title={Mantis: Interleaved multi-image instruction tuning},
  author={Jiang, Dongfu and He, Xuan and Zeng, Huaye and Wei, Cong and Ku, Max and Liu, Qian and Chen, Wenhu},
  journal={arXiv preprint arXiv:2405.01483},
  year={2024}
}

@inproceedings{zelikman2024star,
  title={STaR: Self-taught reasoner bootstrapping reasoning with reasoning},
  author={Zelikman, Eric and Wu, YH and Mu, Jesse and Goodman, Noah D},
  booktitle={Proc. the 36th International Conference on Neural Information Processing Systems},
  volume={1126},
  year={2024}
}

@article{wang2024qwen2,
  title={Qwen2-vl: Enhancing vision-language model's perception of the world at any resolution},
  author={Wang, Peng and Bai, Shuai and Tan, Sinan and Wang, Shijie and Fan, Zhihao and Bai, Jinze and Chen, Keqin and Liu, Xuejing and Wang, Jialin and Ge, Wenbin and others},
  journal={arXiv preprint arXiv:2409.12191},
  year={2024}
}

@article{li2024mini,
  title={Mini-gemini: Mining the potential of multi-modality vision language models},
  author={Li, Yanwei and Zhang, Yuechen and Wang, Chengyao and Zhong, Zhisheng and Chen, Yixin and Chu, Ruihang and Liu, Shaoteng and Jia, Jiaya},
  journal={arXiv preprint arXiv:2403.18814},
  year={2024}
}

@article{suhr2018corpus,
  title={A corpus for reasoning about natural language grounded in photographs},
  author={Suhr, Alane and Zhou, Stephanie and Zhang, Ally and Zhang, Iris and Bai, Huajun and Artzi, Yoav},
  journal={arXiv preprint arXiv:1811.00491},
  year={2018}
}

@article{wu2023q,
  title={Q-bench: A benchmark for general-purpose foundation models on low-level vision},
  author={Wu, Haoning and Zhang, Zicheng and Zhang, Erli and Chen, Chaofeng and Liao, Liang and Wang, Annan and Li, Chunyi and Sun, Wenxiu and Yan, Qiong and Zhai, Guangtao and others},
  journal={arXiv preprint arXiv:2309.14181},
  year={2023}
}

@inproceedings{fu2025blink,
  title={Blink: Multimodal large language models can see but not perceive},
  author={Fu, Xingyu and Hu, Yushi and Li, Bangzheng and Feng, Yu and Wang, Haoyu and Lin, Xudong and Roth, Dan and Smith, Noah A and Ma, Wei-Chiu and Krishna, Ranjay},
  booktitle={European Conference on Computer Vision},
  pages={148--166},
  year={2025},
  organization={Springer}
}

@inproceedings{li2024mvbench,
  title={Mvbench: A comprehensive multi-modal video understanding benchmark},
  author={Li, Kunchang and Wang, Yali and He, Yinan and Li, Yizhuo and Wang, Yi and Liu, Yi and Wang, Zun and Xu, Jilan and Chen, Guo and Luo, Ping and others},
  booktitle={Proceedings of the IEEE/CVF Conference on Computer Vision and Pattern Recognition},
  pages={22195--22206},
  year={2024}
}

@inproceedings{liu2024improved,
  title={Improved baselines with visual instruction tuning},
  author={Liu, Haotian and Li, Chunyuan and Li, Yuheng and Lee, Yong Jae},
  booktitle={Proceedings of the IEEE/CVF Conference on Computer Vision and Pattern Recognition},
  pages={26296--26306},
  year={2024}
}

@article{li2024llava,
  title={Llava-next-interleave: Tackling multi-image, video, and 3d in large multimodal models},
  author={Li, Feng and Zhang, Renrui and Zhang, Hao and Zhang, Yuanhan and Li, Bo and Li, Wei and Ma, Zejun and Li, Chunyuan},
  journal={arXiv preprint arXiv:2407.07895},
  year={2024}
}

@article{liu2024mia,
  title={Mia-dpo: Multi-image augmented direct preference optimization for large vision-language models},
  author={Liu, Ziyu and Zang, Yuhang and Dong, Xiaoyi and Zhang, Pan and Cao, Yuhang and Duan, Haodong and He, Conghui and Xiong, Yuanjun and Lin, Dahua and Wang, Jiaqi},
  journal={arXiv preprint arXiv:2410.17637},
  year={2024}
}

@inproceedings{yue2024mmmu,
  title={Mmmu: A massive multi-discipline multimodal understanding and reasoning benchmark for expert agi},
  author={Yue, Xiang and Ni, Yuansheng and Zhang, Kai and Zheng, Tianyu and Liu, Ruoqi and Zhang, Ge and Stevens, Samuel and Jiang, Dongfu and Ren, Weiming and Sun, Yuxuan and others},
  booktitle={Proceedings of the IEEE/CVF Conference on Computer Vision and Pattern Recognition},
  pages={9556--9567},
  year={2024}
}

@article{chen2024we,
  title={Are We on the Right Way for Evaluating Large Vision-Language Models?},
  author={Chen, Lin and Li, Jinsong and Dong, Xiaoyi and Zhang, Pan and Zang, Yuhang and Chen, Zehui and Duan, Haodong and Wang, Jiaqi and Qiao, Yu and Lin, Dahua and others},
  journal={arXiv preprint arXiv:2403.20330},
  year={2024}
}

@article{lu2022learn,
  title={Learn to explain: Multimodal reasoning via thought chains for science question answering},
  author={Lu, Pan and Mishra, Swaroop and Xia, Tanglin and Qiu, Liang and Chang, Kai-Wei and Zhu, Song-Chun and Tafjord, Oyvind and Clark, Peter and Kalyan, Ashwin},
  journal={Advances in Neural Information Processing Systems},
  volume={35},
  pages={2507--2521},
  year={2022}
}

@article{yin2023survey,
  title={A survey on multimodal large language models},
  author={Yin, Shukang and Fu, Chaoyou and Zhao, Sirui and Li, Ke and Sun, Xing and Xu, Tong and Chen, Enhong},
  journal={arXiv preprint arXiv:2306.13549},
  year={2023}
}

@article{li2023evaluating,
  title={Evaluating object hallucination in large vision-language models},
  author={Li, Yifan and Du, Yifan and Zhou, Kun and Wang, Jinpeng and Zhao, Wayne Xin and Wen, Ji-Rong},
  journal={arXiv preprint arXiv:2305.10355},
  year={2023}
}

@article{guan2023hallusionbench,
  title={HallusionBench: An Advanced Diagnostic Suite for Entangled Language Hallucination and Visual Illusion in Large Vision-Language Models},
  author={Guan, Tianrui and Liu, Fuxiao and Wu, Xiyang and Xian, Ruiqi and Li, Zongxia and Liu, Xiaoyu and Wang, Xijun and Chen, Lichang and Huang, Furong and Yacoob, Yaser and others},
  journal={arXiv preprint arXiv:2310.14566},
  year={2023}
}

@article{liu2024visual,
  title={Visual instruction tuning},
  author={Liu, Haotian and Li, Chunyuan and Wu, Qingyang and Lee, Yong Jae},
  journal={Advances in neural information processing systems},
  volume={36},
  year={2024}
}

@article{wang2023cogvlm,
  title={Cogvlm: Visual expert for pretrained language models},
  author={Wang, Weihan and Lv, Qingsong and Yu, Wenmeng and Hong, Wenyi and Qi, Ji and Wang, Yan and Ji, Junhui and Yang, Zhuoyi and Zhao, Lei and Song, Xixuan and others},
  journal={arXiv preprint arXiv:2311.03079},
  year={2023}
}

@article{snell2024scaling,
  title={Scaling llm test-time compute optimally can be more effective than scaling model parameters},
  author={Snell, Charlie and Lee, Jaehoon and Xu, Kelvin and Kumar, Aviral},
  journal={arXiv preprint arXiv:2408.03314},
  year={2024}
}

@article{prystawski2024think,
  title={Why think step by step? Reasoning emerges from the locality of experience},
  author={Prystawski, Ben and Li, Michael and Goodman, Noah},
  journal={Advances in Neural Information Processing Systems},
  volume={36},
  year={2024}
}

@article{zhu2025internvl3,
  title={Internvl3: Exploring advanced training and test-time recipes for open-source multimodal models},
  author={Zhu, Jinguo and Wang, Weiyun and Chen, Zhe and Liu, Zhaoyang and Ye, Shenglong and Gu, Lixin and Tian, Hao and Duan, Yuchen and Su, Weijie and Shao, Jie and others},
  journal={arXiv preprint arXiv:2504.10479},
  year={2025}
}

@article{Wang2024Qwen2VLEV,
  title={Qwen2-VL: Enhancing Vision-Language Model's Perception of the World at Any Resolution},
  author={Peng Wang and Shuai Bai and Sinan Tan and Shijie Wang and Zhihao Fan and Jinze Bai and Ke-Yang Chen and Xuejing Liu and Jialin Wang and Wenbin Ge and Yang Fan and Kai Dang and Mengfei Du and Xuancheng Ren and Rui Men and Dayiheng Liu and Chang Zhou and Jingren Zhou and Junyang Lin},
  journal={ArXiv},
  year={2024},
  volume={abs/2409.12191},
  url={https://api.semanticscholar.org/CorpusID:272704132}
}

@article{Bai2025Qwen25VLTR,
  title={Qwen2.5-VL Technical Report},
  author={Shuai Bai and Keqin Chen and Xuejing Liu and Jialin Wang and Wenbin Ge and Sibo Song and Kai Dang and Peng Wang and Shijie Wang and Jun Tang and Humen Zhong and Yuanzhi Zhu and Mingkun Yang and Zhaohai Li and Jianqiang Wan and Pengfei Wang and Wei Ding and Zheren Fu and Yiheng Xu and Jiabo Ye and Xi Zhang and Tianbao Xie and Zesen Cheng and Hang Zhang and Zhibo Yang and Haiyang Xu and Junyang Lin},
  journal={ArXiv},
  year={2025},
  volume={abs/2502.13923},
  url={https://api.semanticscholar.org/CorpusID:276449796}
}

@inproceedings{liu2025tfcustom,
  title={TFCustom: Customized Image Generation with Time-Aware Frequency Feature Guidance},
  author={Liu, Mushui and She, Dong and Pang, Jingxuan and Huang, Qihan and Ying, Jiacheng and He, Wanggui and Hou, Yuanlei and Fu, Siming},
  booktitle={CVPR},
  pages={2714--2723},
  year={2025}
}

@article{wang2025mint,
  title={Mint: Multi-modal chain of thought in unified generative models for enhanced image generation},
  author={Wang, Yi and Liu, Mushui and He, Wanggui and Zhang, Longxiang and Huang, Ziwei and Zhang, Guanghao and Shu, Fangxun and Tao, Zhong and She, Dong and Yu, Zhelun and others},
  journal={arXiv preprint arXiv:2503.01298},
  year={2025}
}

@inproceedings{he2025mars,
  title={Mars: Mixture of auto-regressive models for fine-grained text-to-image synthesis},
  author={He, Wanggui and Fu, Siming and Liu, Mushui and Wang, Xierui and Xiao, Wenyi and Shu, Fangxun and Wang, Yi and Zhang, Lei and Yu, Zhelun and Li, Haoyuan and others},
  booktitle={AAAI},
  year={2025}
}

@article{ma2025followcreation,
  title={Follow-Your-Creation: Empowering 4D Creation through Video Inpainting},
  author={Ma, Yue and Feng, Kunyu and Zhang, Xinhua and Liu, Hongyu and Zhang, David Junhao and Xing, Jinbo and Zhang, Yinhan and Yang, Ayden and Wang, Zeyu and Chen, Qifeng},
  journal={arXiv preprint arXiv:2506.04590},
  year={2025}
}

@inproceedings{ma2024followyouremoji,
  title={Follow-your-emoji: Fine-controllable and expressive freestyle portrait animation},
  author={Ma, Yue and Liu, Hongyu and Wang, Hongfa and Pan, Heng and He, Yingqing and Yuan, Junkun and Zeng, Ailing and Cai, Chengfei and Shum, Heung-Yeung and Liu, Wei and others},
  booktitle={SIGGRAPH Asia},
  pages={1--12},
  year={2024}
}

@inproceedings{esser2024scalingrectifiedflowtransformers,
  title={Scaling rectified flow transformers for high-resolution image synthesis},
  author={Esser, Patrick and Kulal, Sumith and Blattmann, Andreas and Entezari, Rahim and M{\"u}ller, Jonas and Saini, Harry and Levi, Yam and Lorenz, Dominik and Sauer, Axel and Boesel, Frederic and others},
  booktitle={ICML},
  year={2024}
}

@inproceedings{feng2025dit4edit,
  title={Dit4edit: Diffusion transformer for image editing},
  author={Feng, Kunyu and Ma, Yue and Wang, Bingyuan and Qi, Chenyang and Chen, Haozhe and Chen, Qifeng and Wang, Zeyu},
  booktitle={AAAI},
  year={2025}
}

\clearpage
\section{Appendix}

\section{More Experimental Results}
\subsection{Fine-tuned results of different models}
\label{fine-tuned}
To further verify that the innovation of our approach is not solely dependent on the dataset, we analyzed the results of different models fine-tuned on the CMMCoT-260k dataset. As shown in \ref{tab:model_performance_compare}, the experimental results show that the LLaVA and Mantis models exhibit a significant performance decline after CMMCoT fine-tuning, likely due to their lack of visual localization capabilities, which is exacerbated by the fine-tuning process. The LLaVA-OV model, which incorporates localization data during pre-training, shows a smaller degree of performance degradation, yet still experiences some decline, possibly due to the absence of explicit localization mechanisms in its output. The Qwen2-VL model, equipped with localization capabilities through pre-training and architectural design, demonstrates moderate improvement after fine-tuning. The results confirm that only when the original multimodal model possesses strong localization abilities can the full potential of the CMMCoT method be realized.

\begin{table*}[ht]
\centering
\caption{Performance comparison across multimodal benchmarks.}
\label{tab:model_performance_compare}
\begin{tabular}{lccccccc}
\toprule
\textbf{Model} & \textbf{Training} & \textbf{BLINK} & \textbf{Mantis} & \textbf{NLVR2} & \textbf{MVBench} & \textbf{Q-Bench} & \textbf{Avg} \\
\midrule 
\rowcolor{gray!15}
LLaVA-v1.5 (7B) & Frozen & 37.1 & 41.9 & 52.1 & 36.0 & 53.9 & 44.2 \\
\rowcolor{gray!15}
LLaVA-v1.5 (7B) & Fine-tuned & 35.5 & 30.4 & 49.0 & 32.7 & 40.2 & 37.6 \\
LLaVA-v1.6 (7B) & Frozen & 39.6 & 45.6 & 58.9 & 40.9 & 58.9 & 48.8 \\
LLaVA-v1.6 (7B) & Fine-tuned & 38.5 & 40.8 & 53.2 & 38.2 & 55.9 & 45.3 \\
\rowcolor{gray!15}
Mantis-Idefics2 (8B) & Frozen & 49.1 & 57.1 & 89.7 & 51.4 & 75.3 & 64.5 \\
\rowcolor{gray!15}
Mantis-Idefics2 (8B) & Fine-tuned & 46.3 & 59.4 & 84.6 & 40.8 & 69.6 & 60.1 \\
LLaVA-OV (7B) & Frozen & 48.2 & 64.2 & 89.4 & 56.7 & 74.5 & 66.6 \\
LLaVA-OV (7B) & Fine-tuned & 46.9 & 62.5 & 87.4 & 50.8 & 73.3 & 64.2 \\
\rowcolor{gray!15}
Qwen2-VL (7B) & Frozen & 51.8 & 64.8 & 84.0 & 50.5 & 71.2 & 64.5 \\
\rowcolor{gray!15}
Qwen2-VL (7B) & Fine-tuned & 52.6 & 66.8 & 88.5 & 50.0 & 73.4 & 66.3 \\
Qwen2.5-VL (7B) & Frozen & 55.3 & 69.8 & 88.3 & 74.7 & 77.7 & 73.2 \\
Qwen2.5-VL (7B) & Fine-tuned & 55.6 & 70.6 & 88.7 & 74.9 & 77.9 & 73.5 \\
\midrule
\rowcolor{gray!15}
Qwen2.5-VL(7B) & \textbf{Our Method} & \textbf{56.8} & \textbf{72.2} & \textbf{89.9} & \textbf{75.8} & \textbf{78.5} & \textbf{74.6} \\
\bottomrule
\end{tabular}
\end{table*}

\section{Details of CMMCoT-260K} 
\label{details}
CMMCoT-260K encompasses four distinct types of data: {Caption}, {Co-reference}, {Comparison}, and {Reason}.

\paragraph{Caption} The caption data type is employed to generate natural language descriptions for images, serving as a direct expression of the semantic information contained within the image.
It provides the model with the ability to comprehend and semantically parse image content, facilitating tasks such as annotating and generating descriptions for multiple images.

\paragraph{Co-reference} Co-reference data involves identifying and linking elements that refer to the same entity across different images. This data type assists the model in understanding entity consistency within a single image or among multiple images.
It enhances the model's ability to maintain consistent entity reasoning across multiple images, thereby strengthening its understanding of complex scenes.

\paragraph{Comparison} The comparison data type is utilized for conducting comparative analyses between multiple images, identifying similarities and differences among elements in different images. It improves the model's comparative and contrastive analytical capabilities, enabling it to make more nuanced and accurate judgments in tasks involving multiple images, such as in visual question answering.

\paragraph{Reason} The reason data type pertains to more complex reasoning tasks, requiring the model to perform logical reasoning and decision-making based on relationships among multiple images. This typically necessitates contextual understanding and causal relationship analysis between images. It enhances the model's capacity to solve advanced reasoning problems, making it applicable for understanding more complex task contexts.

\begin{table*}[htbp]
  \centering
  \caption{The statistics of CMMCoT-260K}
  \begin{tabular}{lllc}
    \toprule
    \textbf{Skill} & \textbf{Source} & \textbf{Dataset} & \textbf{Instances} \\
    \midrule
    \rowcolor{gray!10}
    Caption & grit & caption\_grit-dual\_20k & 20k \\
    \rowcolor{gray!10}
            & grit & caption\_grit-triple\_20k & 20k \\
    \rowcolor{gray!10}
            & grit & caption\_grit-quad\_10k & 10k \\
    \midrule
    Co-reference & flickr30k-entities& coreference\_shikra-dual\_10k & 10k \\
                 & vocot & coreference\_vocot-dual\_50k & 40k \\
                 & vocot & coreference\_vocot-triple\_30k & 30k \\
                 & vocot & coreference\_vocot-quad\_10k & 10k \\
    \midrule
    \rowcolor{gray!10}
    Comparison & mantis & birds-to-words& 3k \\
    \rowcolor{gray!10}
               & mantis & dreamsim& 15k \\
    \midrule
    Reason & mantis & nlrv2& 85k \\
           & mantis & imagecode& 17k \\
    \midrule
    \rowcolor{gray!20}
    \textbf{Total} & & & \textbf{260k} \\
    \bottomrule
  \end{tabular}
\end{table*}

\newpage
\section{More Examples}
\label{cases}
\begin{figure*}[ht]
    \centering
    \includegraphics[width=1\linewidth]{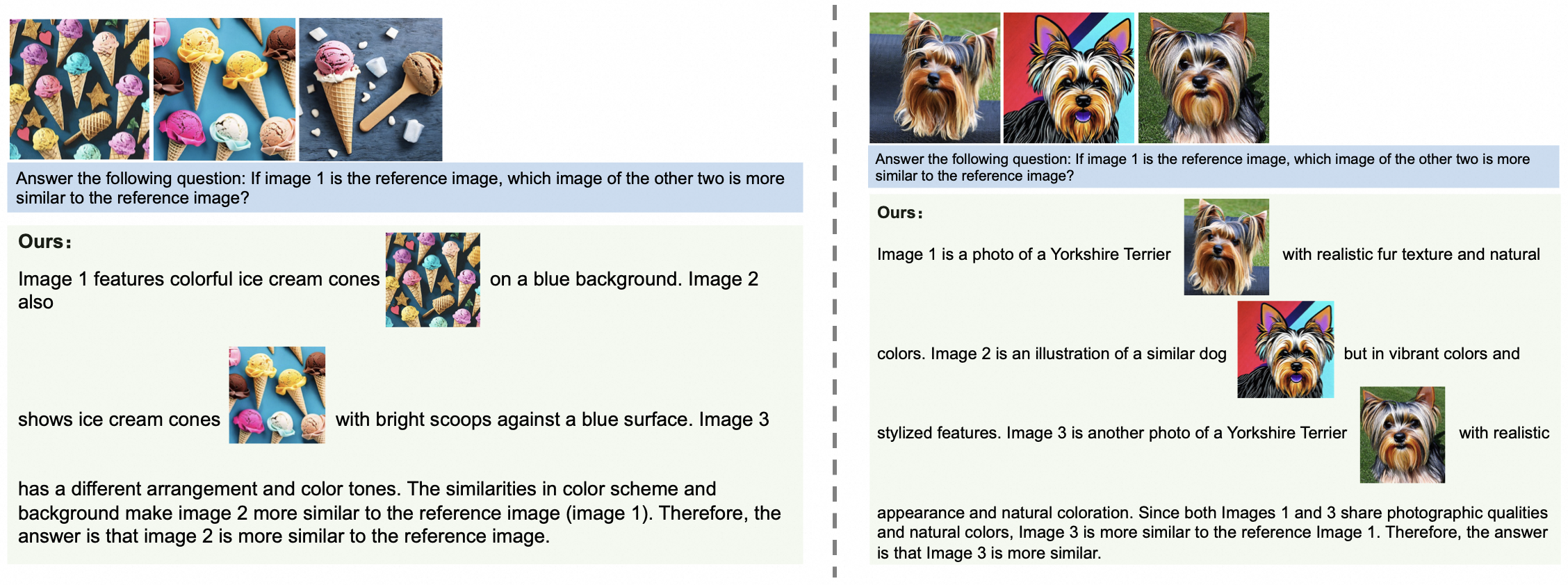}
    \caption{Visualization results of our CMMCoT task to illustrate the fine-grained reasoning ability of our model.}
    \label{fig:demo7}
\end{figure*}

\begin{figure*}[ht]
    \centering
    \includegraphics[width=1\linewidth]{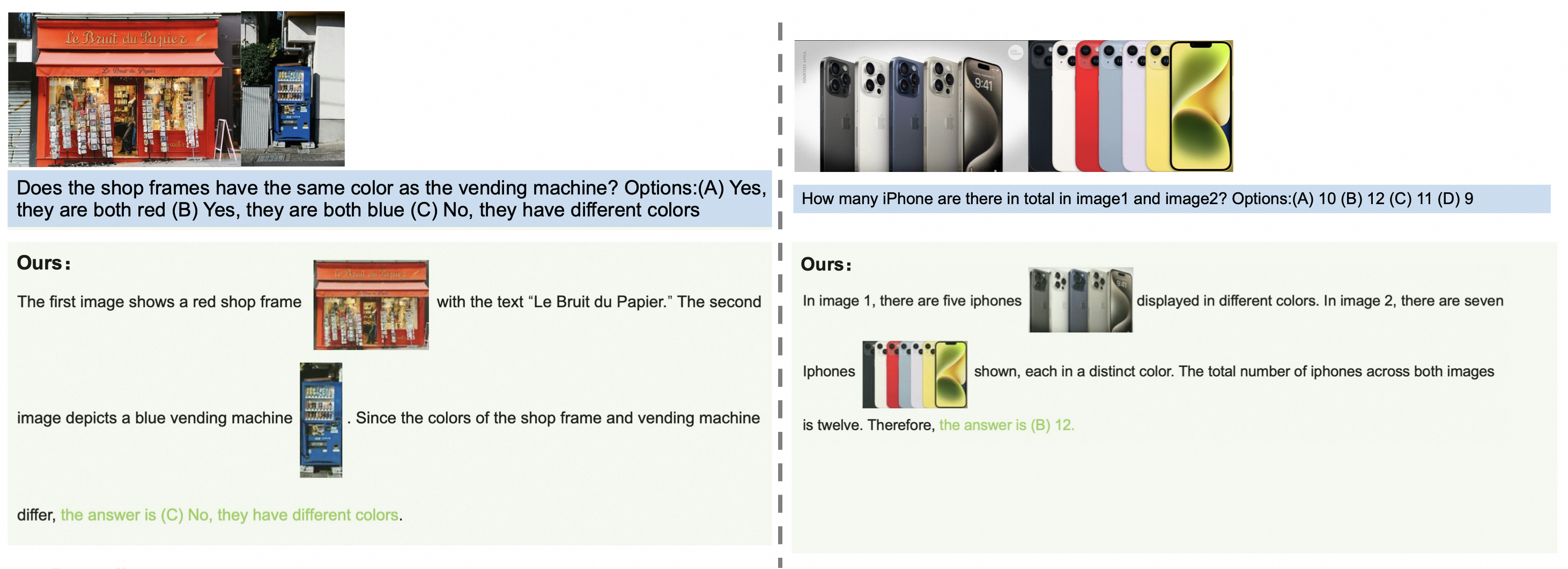}
    \caption{More examples of multi-image tasks.}
    
    \label{fig:demo8}
\end{figure*}

\begin{figure*}[ht]
    \centering
    \includegraphics[width=1\linewidth]{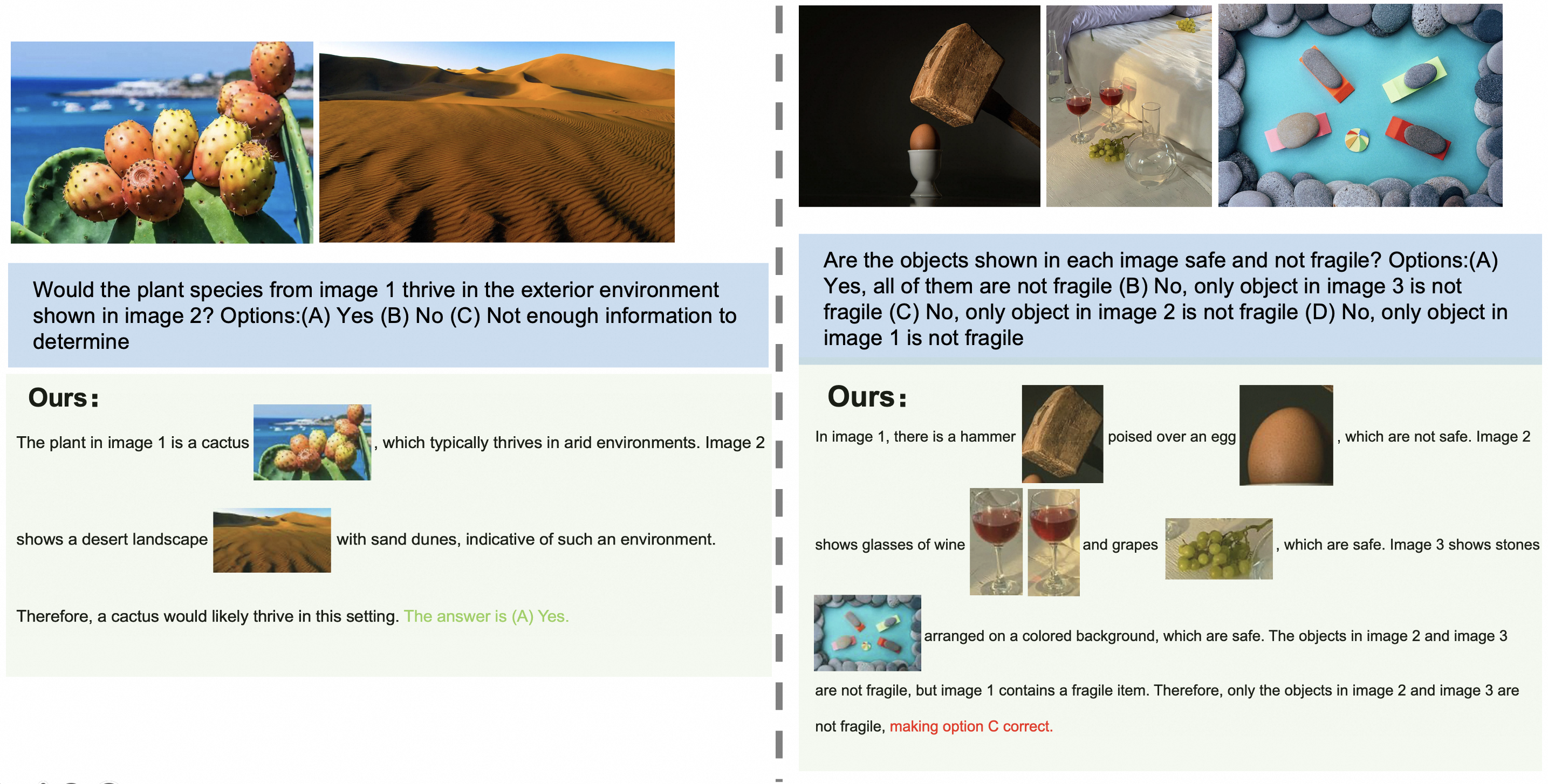}
    \caption{More examples of multi-image tasks.}
    
    \label{fig:demo9}
\end{figure*}

We also provide additional visualization examples, as shown in Figure \ref{fig:demo7}, Figure \ref{fig:demo8} and Figure \ref{fig:demo9}. These successful cases further demonstrate the efficacy of our model. From the bad case in Figure \ref{fig:demo9}, we can observe that despite the model making an incorrect prediction, it is still capable of extracting and analyzing key visual elements mentioned in the text.


\end{document}